\documentclass[twoside,11pt]{article}

\usepackage{blindtext}

\usepackage[preprint]{dmlr2e_pre}

\usepackage{multirow}
\usepackage{algorithm}
\usepackage{amsmath}
\usepackage{algpseudocode}
\usepackage[T1]{fontenc}
\usepackage[utf8]{inputenc}
\usepackage{orcidlink}

\usepackage{lastpage}
\def\openreview{\url{ }} 

\ShortHeadings{Global and Dense Embeddings of Earth}{Major TOM Floating in the Latent Space}
\firstpageno{1}

\begin{document}

\title{Global and Dense Embeddings of Earth:\\Major TOM Floating in the Latent Space}

\author{\name Mikolaj Czerkawski* \orcidlink{0000-0002-0927-0416} \email mikolaj.czerkawski@esa.int \\
       \addr $\Phi$-lab, European Space Agency\\
       Frascati, Italy
       \AND
       \name Marcin Kluczek* \orcidlink{0000-0003-2133-0984} \email mkluczek@cloudferro.com \\
       \addr CloudFerro\\
       Warsaw, Poland
       \AND
       \name  J\k{e}drzej S. Bojanowski \orcidlink{0000-0001-8460-4183} \email jbojanowski@cloudferro.com  \\
       \addr CloudFerro\\
       Warsaw, Poland
       \AND
       \addr *Equal Contribution
       }


\maketitle

\begin{abstract}

    With the ever-increasing volumes of the Earth observation data present in the archives of large programmes such as Copernicus, there is a growing need for efficient vector representations of the underlying raw data. The approach of extracting feature representations from pretrained deep neural networks is a powerful approach that can provide semantic abstractions of the input data. However, the way this is done for imagery archives containing geospatial data has not yet been defined. In this work, an extension is proposed to an existing community project, Major TOM, focused on the provision and standardization of open and free AI-ready datasets for Earth observation. Furthermore, four global and dense embedding datasets are released openly and for free along with the publication of this manuscript, resulting in the most comprehensive global open dataset of geospatial visual embeddings in terms of covered Earth's surface.

\end{abstract}


\section{Introduction}

    There is one trend in the domain of information technologies that does seem more consistent than anything else in recent decades, and it is the growth of produced data. Whether it is a piece of information generated online or a series of bytes transmitted by a satellite, the rate at which new data are produced keeps increasing every year.

    While every new piece of data can be considered valuable and beneficial, the increasing volume makes large-scale analyses more challenging and computationally expensive. This phenomenon is already being observed in the domain of Earth observation data, where every single year, petabytes of new data captured by the satellites enter the archives. For that reason, it is crucial to develop methods capable of navigating vast databases of imagery at speed and low computational costs.

    \begin{figure}
        \centering
        \includegraphics[width=1.0\linewidth]{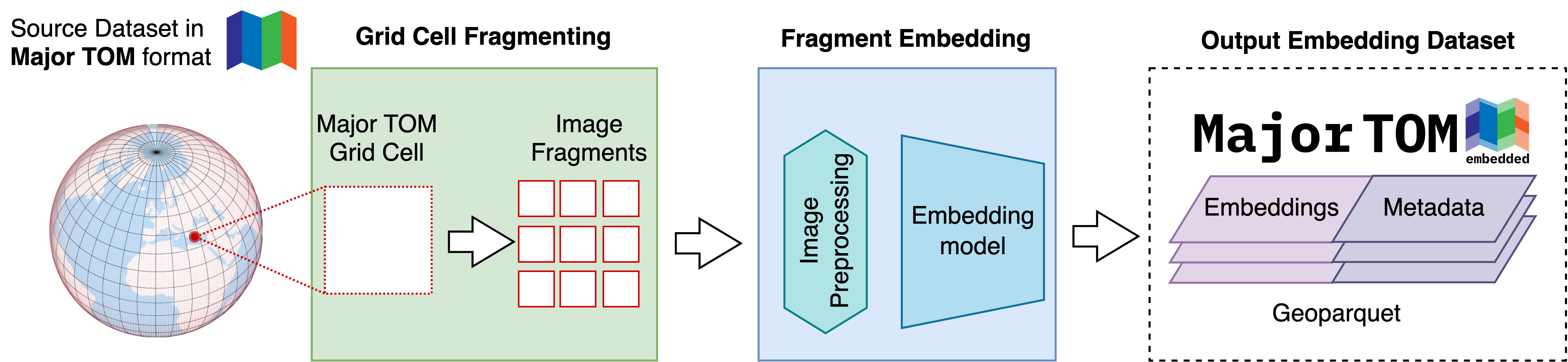}
        \caption{The pipeline building Major TOM embedding expansions according to the proposed standard. It begins with grid cell fragmenting, image preprocessing, and embedding and packing into the geoparquet archive format.}
        \label{fig:intro_diagram}
    \end{figure}

    A solution to this challenge can come in the form of extracting semantic features from the visual data at a large scale, in a process known as embedding~\citep{devise_visual_embeddings,WordinVectorSpace}. While manually engineered features could be used to compress the images into efficient vector representations, this approach has predominantly become effective with the development of large pre-trained deep neural networks, which can learn powerful representations of the input data. This includes methods, which rely solely on self-supervision.

    The process of embedding data into a compressed representation space is far from new and has been explored for over a decade~\citep{WordinVectorSpace,devise_visual_embeddings}, accelerated by the growth in popularity of deep learning pipelines and models~\citep{gordoImageSearch}. Visual descriptors, which play a similar role to embeddings (and sometimes could even be synonymous) have also been a popular topic for many years in computer vision~\citep{Visualquery,fastspatialmatching,SpatiallyConstrainedSimilarity,ImageSearchwithshortVectors}.
    
    In the field of Earth observation, several works have explored the use of vector representations for geospatial data. Most importantly, this includes work that attempted to learn meaningful representations of geospatial coordinates~\citep{yin2019-gps2vec,yin2021-gps2vec+,EmbeddingEarth,satclip}. However, most of these efforts focus on learning a new latent space rather than applying an existing embedding function at a large scale. In this direction, LAION-EO~\citep{LAION_EO} is a notable work, where 5 billion of images coming from the web (LAION-5B dataset~\citep{laion5b}) have been embedded with a CLIP model~\citep{clip} and used to retrieve instances of satellite imagery from billions of unorganized internet data.

    This work attempts to explore the direction of embedding Earth observation images at the global scale with several types of existing models and propose a standard for doing so. For that reason, it largely builds on top of an ongoing community project of Major TOM~\citep{majortom}, which aims to standardize the curation of large Earth observation datasets and set up paths for community-based dataset growth. This work is one example of such a growth, and just like the first Major TOM release, it aims at easy and free access to the datasets and accompanying tools for interacting with them.

    The definition of a standard for embedding expansions brings several benefits. A clear definition leads to reproducible work and embedding databases that can be analyzed, debugged, and fixed if needed. It also ensures that the embedding datasets produced by independent entities are compatible with each other, at least to some degree. It also makes the evaluation of the models used for producing the embeddings more accessible. By precomputing the latent features from a model, the embedding dataset producer removes the larger part of the computational burden, making it possible to evaluate the output representations for various models at a low cost. Finally, by building on top of the Major TOM grid system, embeddings generated according to the standard can be used as aligned embedding time-series data with no further adjustments since the sample footprints are already aligned.

    This work involves computation and analysis of global embeddings from 4 different pre-trained models. This is done in order to highlight the vastly different behaviors, strengths, and weaknesses. It is unlikely that the space of Earth observation will be dominated by a single general-purpose model. At the moment, the ecosystem of AI for satellite imagery already involves a large number of models to choose from, operating on various formatting (such as normalization, numbers of bands, processing level) of data, even within a domain of a single sensor. For that reason, an organized practice of inter-comparison between different embedding model candidates is going to be crucial.

    In the first release of Major TOM embeddings, over 169 million embeddings are released as a result of processing more than 62 TB of raw data, encompassing over 3.5 million unique images. The dataset distills information from approximately 9.368 trillion pixels of source imagery. At the time of release, this is the first open dataset of Copernicus embeddings built at this scale, with dense and global coverage across the full acquisition area of the sensor. The remainder of this paper describes the embedding methodology, the release standard, the fragmenting approach, the models selected for this release, and finally, some early visualizations of the global embedding data.

\section{Embedding Methodology}

    The global embedding dataset is built on top of the Major TOM Core datasets~\citep{majortom}, which enable fast and free access to more than 60 TB of AI-ready Copernicus data, densely covering the globe with multiple sensing modalities and processing levels. In this release, three core datasets were used: \href{https://huggingface.co/datasets/Major-TOM/Core-S2L1C}{\texttt{Major-TOM/Core-S2L1C}}, \href{https://huggingface.co/datasets/Major-TOM/Core-S2L2A}{\texttt{Major-TOM/Core-S2L2A}}, and \href{https://huggingface.co/datasets/Major-TOM/Core-S1RTC}{\texttt{Major-TOM/Core-S1RTC}}, corresponding to Sentinel-2 L1C and L2A, as well as Sentinel-1 RTC products.

    The images in those datasets cover an area of 10.68 by 10.68 square kilometers (containing a full Major TOM grid cell and some margin), which corresponds to 1,068 by 1,068 pixels in the resolution of 10 meters. Since the majority of pre-trained deep neural networks operate on smaller input shapes (such as 224 pixels), the original images from Major TOM had to go through a fragmenting stage. It is worth to note that many model interfaces automatically resize input images of varying shapes to fit the supported input dimensions. However, for over-sized input images, this process inevitably results in information loss. Such a loss could be particularly undesirable for remote sensing data, which often contains fine details crucial for analysis.

    The complete pipeline starts with the fragmenting function as shown in Figure~\ref{fig:intro_diagram}, then proceeds to the model-specific image preprocessing (which also includes the appropriate normalization function) before the data is fed to the embedding model. Finally, the output embedding and metadata (footprint geometries, pixel boundary box, source satellite product information, etc.) are arranged into geoparquet archives, which constitute the output embedding dataset.

    In general, the normalization follows the specific approach associated with a given embedding model. For the general-purpose vision models which have been trained on standard image data (generally in the range of 0 to 1 and with values stored in 8-bit integers), special steps were taken to map the RGB channels of the multi-spectral Sentinel-2 data to a scaling, which resembles the true color. This was done by multiplying the reflectance value by a factor of 2.5 and then clipping the image values between 0 and 1.
    
    In summary, this release delivers embeddings of 4 models listed in Table~\ref{tab:model_summary}. The following sections provide more detail about the data format (Section~\ref{sec:release_standard}), the fragmenting operation (Section~\ref{sec:fragmenting}) and the process of model selection (Section~\ref{sec:models}).
    
    \begin{table}
        \centering
            \caption{Stastistics of the released datasets: \textit{Model} - model used for the embedding, \textit{Source} - source Major TOM Core dataset, \textit{Embeddings} - total number of output embeddings, \textit{Grid cells} - total number of processed Major TOM grid cells, $T_e$ - grid cell processing time on 1$\times$L40S, $T_T$ - total execution time (on 2$\times$L40S)}
                \begin{tabular}{|p{2.5cm}|p{1.1cm}|c|c|c|c|c|} \hline 
                     Model & Source &Patch size&  Embeddings &  Grid cells &  $T_e$ & $T_T$\\ \hline 
                     \href{https://huggingface.co/datasets/Major-TOM/Core-S2L1C-SSL4EO}{SSL4EO-S2} \citep{ssl4eo} & S2L1C &224 x 224&  56,147,150&  2,245,886&  0.41 s& 125 h\\ \hline 
                     \href{https://huggingface.co/datasets/Major-TOM/Core-S1RTC-SSL4EO}{SSL4EO-S1} \citep{ssl4eo} & S1RTC &224 x 224&  36,748,875&  1,469,955&  3.34 s& 60 h\\ \hline 
                     \href{https://huggingface.co/datasets/Major-TOM/Core-S2RGB-SigLIP}{SigLIP(RGB)} \citep{siglip} &S2L2A (RGB) &384 x 384&  20,212,974&  2,245,886&  0.12 s& 111 h\\ \hline 
                     \href{https://huggingface.co/datasets/Major-TOM/Core-S2RGB-DINOv2}{DINOv2(RGB)} \citep{dinov2}  &S2L2A (RGB) &224 x 224&  56,147,150&  2,245,886&  0.33 s& 37 h\\ \hline
                \end{tabular}
        \label{tab:model_summary}
    \end{table}

\section{Release Standard: Major TOM Embedding Expansions}
\label{sec:release_standard}

    The quality of the output release standard is a key factor in the context of usability and reproducibility of the embedding datasets. This relates to both the format of the metadata as well as the embeddings themselves. In this release, the embeddings have been combined with the metadata into joint archive files in the Parquet format, which allows for reading of isolated columns of data and provides built-in compression of the columns. Storing both embeddings and their metadata in this combined form is a step aimed at reducing the risk of potential bugs when the embedding data needs to be matched with its context information from metadata.

    \subsection{Metadata}

        Table~\ref{tab:geoparquet_columns} outlines the fields of the geoparquet output file in the Major TOM Embedding expansions.  This includes embeddings (stored as a numpy array object), and a set of metadata to ensure reproducibility as well as transparency of the dataset. The \texttt{unique\_id} column is a checksum calculated based on the \texttt{geometry}, \texttt{timestamp}, \texttt{product\_id} and the \texttt{embedding} for each processed fragment to improve data integrity.

        The spatial context of the Major TOM grid cell is described by the \texttt{grid\_cell}, \texttt{grid\_row\_u}, and \texttt{grid\_col\_r} (the integer index data provided to enable fast filtering), along with the satellite product information in \texttt{product\_id} and the corresponding \texttt{timestamp}.

        Finally, there is a set of columns designed to provide the context of the individual fragment used as input for the embedding model. This is described by the \texttt{geometry} column (in the WGS84 coordinate reference system), \texttt{utm\_footprint} (stored as well-known text representation of geometry), and \texttt{utm\_crs}. For fast indexing of the original Major TOM sample, \texttt{pixel\_bbox} defines the exact indices used to extract a fragment from the original image. Finally, \texttt{centre\_lat} and \texttt{centre\_lon} are provided as floats representing the latitude and longitude of the fragments to enable fast querying and filtering.

        \begin{table}[]
            \centering
            \caption{Columns of the Major TOM embedding Geoparquet file}
            \begin{tabular}{|l|l|p{9cm}|}
                \hline
                Field& Type & Description \\
                \hline
                \texttt{unique\_id} & string & hash generated from \texttt{geometry}, \texttt{time}, \texttt{product\_id} and the \texttt{embedding} \\
                \texttt{embedding} & array & raw embedding array \\
                \hline
                \texttt{grid\_cell} & string & Major TOM cell\\
                \texttt{grid\_row\_u} & int & Major TOM cell row \\
                \texttt{grid\_col\_r} & int & Major TOM cell col\\
                \texttt{product\_id} & string & ID of the original product\\
                \texttt{timestamp} & string & Timestamp of the sample\\
                \hline
                \texttt{geometry} & geometry & Polygon footprint (WGS84) of the fragment \\
                \texttt{utm\_footprint} & string & Polygon footprint (image UTM) of the fragment \\
                \texttt{utm\_crs} & & CRS of the original product\\
                \texttt{pixel\_bbox} & bbox & Boundary box of the fragment (pixels) \\               
                \texttt{centre\_lat} & float & Centre of the fragment latitude\\
                \texttt{centre\_lon} & float & Centre of the fragment longitude\\
                \hline
            \end{tabular}
            \label{tab:geoparquet_columns}
        \end{table}

    \subsection{Embedding Data Format}

        The GeoParquet format is an extension to the standard column-based Parquet file format, which most importantly introduces support for a geometry-type column. This way the files can benefit from the efficiency and convenience of the Parquet standard while being interoperable with standard geospatial data processing pipelines. This choice is also compatible with the original Major TOM Core datasets which are currently stored in regular parquet files.

        The embeddings themselves are stored as a numpy array in one of the columns. While Parquet supports byte arrays in columns (in fact, this is how GeoTIFF data is stored in the original Major TOM Core dataset parquets), the bytes require a specific interpretation context. For example, some of the embedding datasets might choose to store the data in higher or lower precision of number representation than the common 4-byte floating point. In case it happens, the user of the dataset must be fully aware of this change in order to read raw bytes, and mistakes can happen. For that reason, the current approach is to store the numpy objects which are more heavy-weight but provide the necessary context of the data type, removing the risk of improper reading of the array.

        This choice does not exclude the option of working with raw byte types. In fact, the software release demonstrates how to turn a set of Major TOM Embedding dataset parquets into a set of metadata and a byte array format to improve reading speed. However, this process is done from scratch locally, which ensures that the data is first downloaded in a combined GeoParquet format, ensuring the initial integrity of the archive.

    \begin{figure}[ht]
        \centering
        \includegraphics[width=0.9\linewidth]{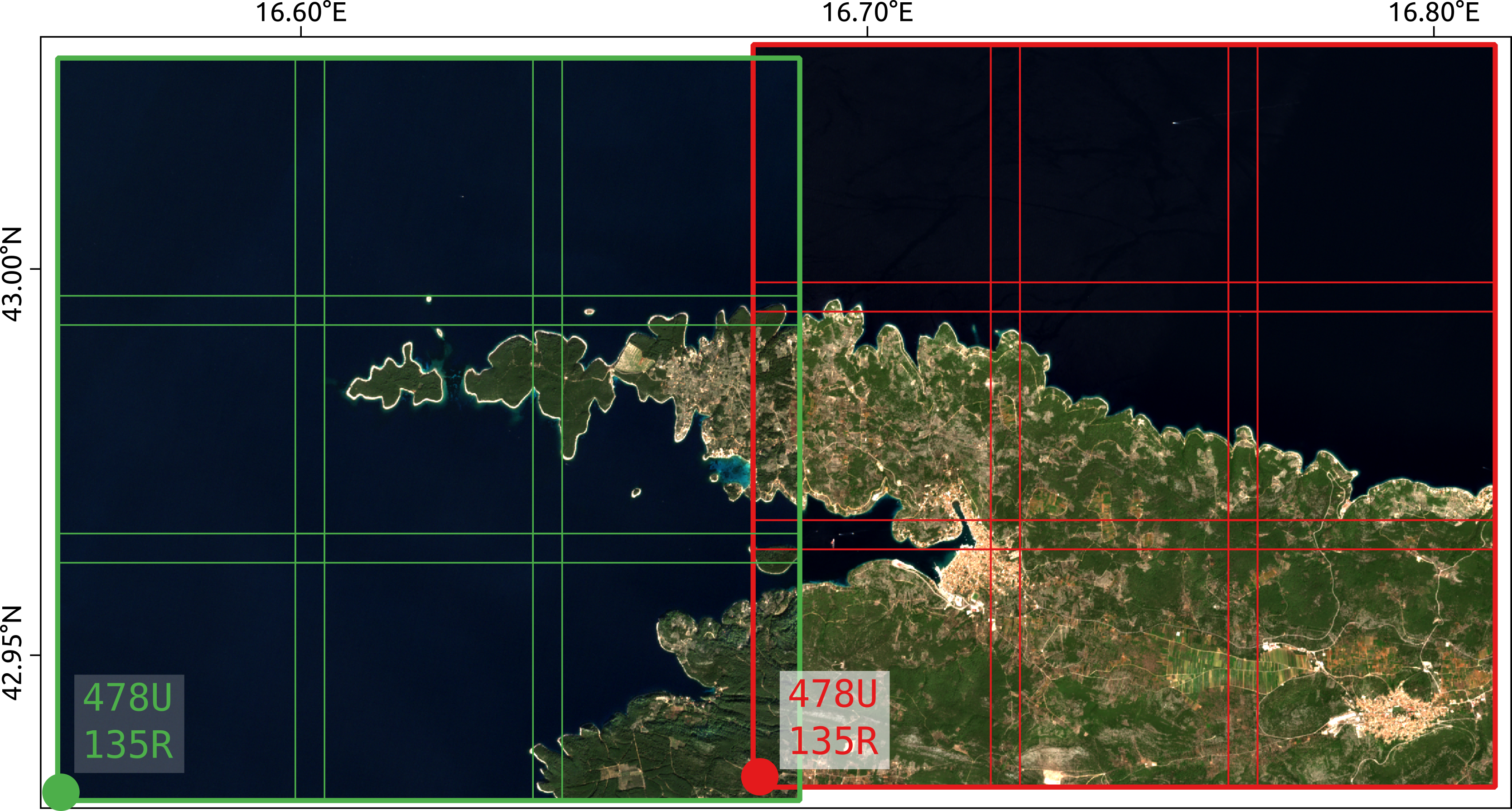}
        \caption{Fragmenting function for SigLIP (fragments of 384 pixels) for 2 independent Major TOM grid cells plotted next to each other. Note that these cells (green and red) are fragmented and processed independently, and are plotted here together for visualisation.}
        \label{fig:cells_siglip}
    \end{figure}

    \begin{figure}[ht]
        \centering
        \includegraphics[width=0.9\linewidth]{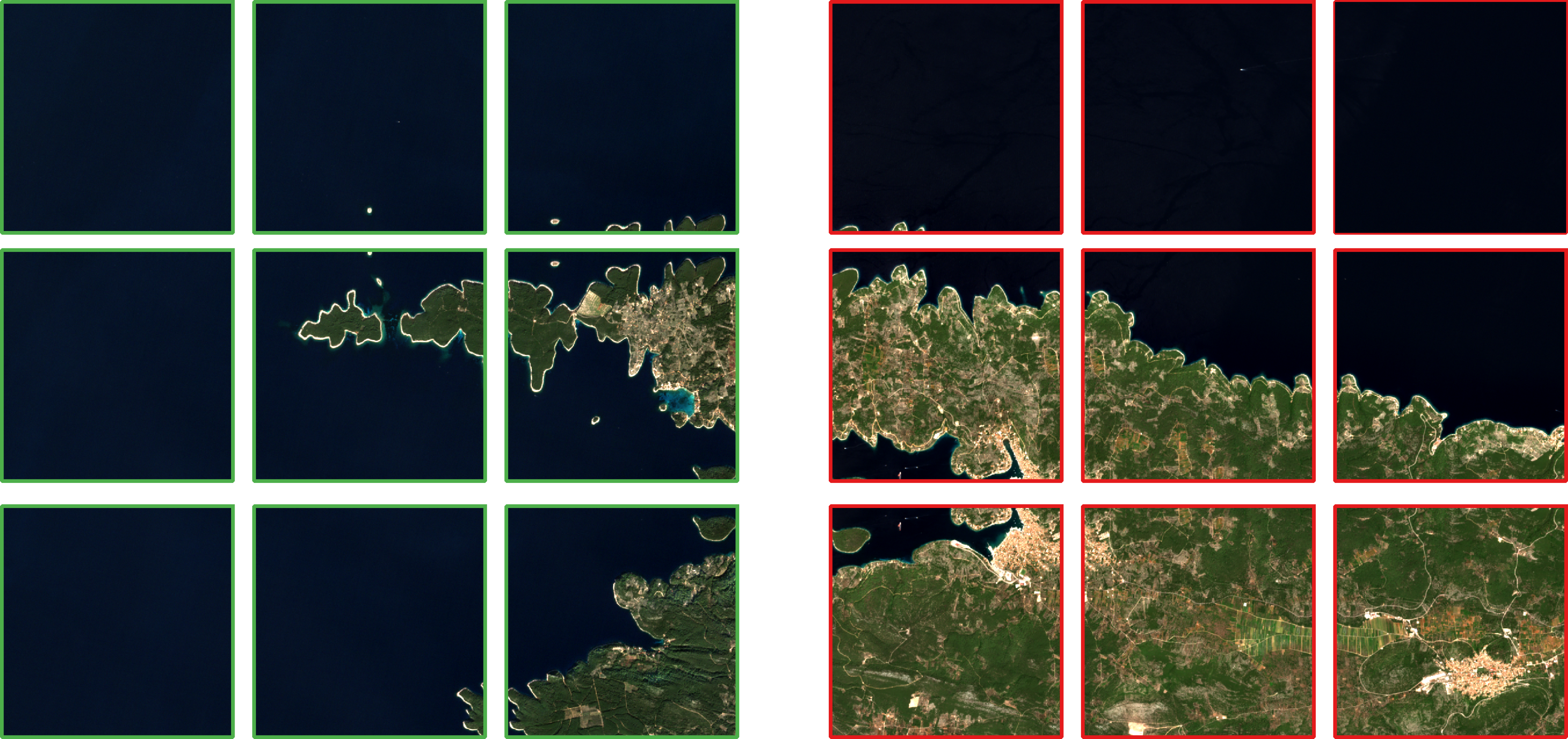}
        \caption{Individual fragments for 2 independent Major TOM grid cells plotted next to each other.}
        \label{fig:fragments_siglip}
    \end{figure}

\section{Fragmenting Major TOM Cells}
\label{sec:fragmenting}

    Just like the original Major TOM datasets were designed with standardization and alignment in mind, the same approach is followed for the embedding expansions. For consistent Major TOM datasets, the footprints of samples covering the same Major TOM grid cell are going to align as long as the original products are in the same projection. While this approach has been designed to make various image datasets compatible with each other, it can also have positive impacts in the context of embedding these datasets.

    Namely, aligned embedding footprints will lead to the emergence of consistent embedding time series. For that reason, this work proposes an algorithm that fragments a given Major TOM cell in a consistent manner and defines a specific configuration. The algorithm configuration is based on an input image size $s_I$ (in the case of Major TOM Sentinel-2 samples, this is 1,068 pixels), target fragment size $s_f$ (which corresponds to the input image size of a given model, such as common 384 pixels), and target overlap $o_t$ fraction (which defines the approximate overlap of consecutive fragments within the cell).

    The algorithm will prioritize the inclusion of all pixels from the source cell (as shown in Figures~\ref{fig:cells_siglip} and~\ref{fig:fragments_siglip}), which is why the overlap $o_t$ will be adjusted in order to cover the complete source image fully.

    \subsection{Algorithm}

    The Algorithm~\ref{alg:frag_fn} contains a complete representation of the fragmenting function, which includes the adjustment of the feasible overlap $s_o$ based on the target overlap fraction $o_t$ and based on that, computation of the total number of fragments $n$ per dimension. It is then followed by a for loop iterating over all fragment indices and computing the row and column offset for a given fragment. These offsets are then used to extract a corresponding crop from the source image $I$ along with the boundary box index. If a fragment is near the far edge of the image and the \texttt{border\_shift} option is set as True, then it will be ensured that the fragment touches the far border of the image, which is done to prevent any source pixel omission (although this effect would be minimal since the target overlap is adjusted based on the image size).

    \begin{algorithm}[ht]
        \caption{Image Fragmentation Algorithm with overlap and border shift parameters}\label{alg:frag_fn}
        \begin{algorithmic}[1]
            \Require Image $I$ of size $s_I \times s_I$, fragment size $s_f$, target overlap $o_t \in [0, 1]$, border\_shift $\in \{\text{False, True}\}$
            \Ensure $s_I \geq s_f$
    
            \If{$s_I = s_f$}
                \State \texttt{fragments[0, 0] = I} \Comment{Single fragment corresponds to the full image.}
            \Else
                \State $s_o \gets s_f \cdot o_t$ \Comment{Target overlap in pixels}
                \State $n \gets \max\left(1, \text{round}\left(\frac{s_I - s_f}{s_f - s_o}\right)\right)$ \Comment{Number of fragments along each axis (excluding border shift)}
                \State $s_o \gets \lceil s_f - \frac{s_I - s_f}{n} \rceil$ \Comment{Adjusted overlap to ensure proper fragment alignment}
    
                \For{$\texttt{row\_idx} = 0, 1, \dots, n-1$} \Comment{Iterate through rows}
                    \For{$\texttt{col\_idx} = 0, 1, \dots, n-1$} \Comment{Iterate through columns}
                        \If{$\texttt{row\_idx} = n-1$ \textbf{and} border\_shift} 
                            \State $\texttt{row\_offset} \gets s_I - s_f$ 
                        \Else
                            \State $\texttt{row\_offset} \gets \texttt{row\_idx} \cdot (s_f - s_o)$ 
                        \EndIf
                        \\
                        \If{$\texttt{col\_idx} = n-1$ \textbf{and} border\_shift} 
                            \State $\texttt{col\_offset} \gets s_I - s_f$ 
                        \Else
                            \State $\texttt{col\_offset} \gets \texttt{col\_idx} \cdot (s_f - s_o)$ 
                        \EndIf
                        \\
                        \State \texttt{xys[row\_idx, col\_idx] = \{\texttt{row\_offset}, \texttt{col\_offset}\}} 
                        \State \texttt{fragments[row\_idx, col\_idx] = I[\texttt{row\_offset}:\texttt{row\_offset}+$s_f$,\\ \hspace{20em}\texttt{col\_offset}:\texttt{col\_offset}+$s_f$]} 
                    \EndFor
                \EndFor
            \EndIf
    
            \State \textbf{Return} fragments, xys
        \end{algorithmic}
    \end{algorithm}

\section{Embedding Models}
\label{sec:models}

    There are quite a few open models trained on large-scale Copernicus data (albeit not as large as Major TOM at the time of writing)~\citep{seco,satmae,satlas,prithvi,dofa,presto} or other types of satellite sensor data~\citep{scale_mae}. The majority of these models operate on Sentinel-2 data, given the versatility of this optical sensor. However, the domain of Sentinel-1 general-purpose models is also growing, at a slower pace though. 

    \subsection{Sentinel-2}

        For Sentinel-2, there is a range of models trained in a self-supervised manner directly on the multi-spectral Level 1C (top of atmosphere) representation~\citep{ssl4eo,decur}. In some cases, like in SatCLIP~\citep{satclip}, these models have been used with Level 2A data (bottom of the atmosphere) by filling in the missing cirrus band with zeroes. There are other pre-trained models which may also be used for this purpose, however, SSL4EO has been prioritised due to the application in independent works (SatCLIP), simple and small (25 million parameters) architecture of ResNet50, easily accessible code, well-known training data, high benchmark performance~\citep{prithvi2}, and inclusion in the torchgeo package (which can lead to higher reproducibility of the approach).

        Another route to obtaining embeddings of Sentinel-2 images is to re-use large pre-trained models from general-purpose computer vision contexts, such as CLIP~\citep{clip}, SigLIP~\citep{siglip}, or DINOv2~\citep{dinov2}. This immediately introduces certain limitations, since only the three RGB can be used, and most of the models will expect images scaled between 0-1 with intensity distribution similar to those observed in the training set. This means that some of the high intensity values detected by the Sentinel-2 sensor must be discarded in the process.

        Finally, most of the mentioned models compute a single global embedding, which represents the input image. A recent model trained with multi-pretext masking self-supervision~\citep{mmearth} can be used to obtain a spatially distributed map of embeddings per image and this model is currently being investigated.

        The models considered in this project are listed in Table~\ref{tab:s2_models}.

        \begin{table}[ht]
            \centering
            \caption{Embeddings of the Sentinel-2 Core dataset}
            \begin{tabular}{|p{3cm}|p{2cm}|p{4cm}|p{4.2cm}|}
                \hline
                Model & Architecture & Notes & Access \\
                \hline
                \multicolumn{4}{|c|}{Sentinel-2 L1C (Top of Atmosphere)}\\
                \hline
                 SSL4EO-S2 \citep{ssl4eo}& ResNet50 & Models trained with various SSL methods on multi-spectral S2 data (L1C). & \href{https://huggingface.co/datasets/Major-TOM/Core-S2L2A-SSL4EO}{hf.co:Major-TOM/Core-S2L2A-SSL4EO} \\
                 \hline
                 DeCUR \citep{decur}& ResNet50 & Multi-modal self-supervision by learning inter- and intra-modal embeddings & (To be computed) \\
                 \hline
                 \multicolumn{4}{|c|}{Sentinel-2 L2A (Bottom of Atmosphere)}\\
                 \hline
                 SigLIP(RGB) \citep{siglip} & ViT & Powerful vision-language model, yet not designed for EO data.& \href{https://huggingface.co/datasets/Major-TOM/Core-S2RGB-SigLIP}{hf.co:Major-TOM/Core-S2RGB-SigLIP} \\
                 \hline
                 DINOv2(RGB) \citep{dinov2} & ViT & Powerful vision model trained with self-supervision. & \href{https://huggingface.co/datasets/Major-TOM/Core-S2RGB-DINOv2}{hf.co:Major-TOM/Core-S2RGB-DINOv2} \\
                 \hline
                 MMEarth \citep{mmearth} & ConvNeXt V2 & Multi-pretext masked autoencoder & (To be computed) \\
                 \hline
            \end{tabular}
            \label{tab:s2_models}
        \end{table}

    \subsection{Sentinel-1}

        There do not seem to be as many models available for Sentinel-1 as for Sentinel-2 openly available, as indicated by the presence of only four pre-trained model weights on the torchgeo website~\citep{torchgeo}. These 4 models include SSL4EO-S12~\citep{ssl4eo}, DeCUR~\citep{decur}, and SATLAS~\citep{satlas}. As mentioned earlier, due to the same reasons for preference towards the SSL4EO models, this one was prioritised for the first release of Sentinel-1 embeddings, as shown in Table~\ref{tab:s1_models}.

        \begin{table}[ht]
            \centering
            \caption{Embeddings of the Sentinel-1 Core dataset}
            \begin{tabular}{|l|l|l|}
                \hline
                Model & Architecture & Access \\
                \hline
                 \multicolumn{3}{|c|}{Sentinel-1 RTC (Radiometrically Terrain Corrected )}\\
                \hline
                 SSL4EO-S1 & ResNet50 & \href{https://huggingface.co/datasets/Major-TOM/Core-S1RTC-SSL4EO}{hf.co:Major-TOM/Core-S1RTC-SSL4EO} \\
                 DeCUR & ResNet50 & (To be computed) \\
                 \hline
            \end{tabular}
            \label{tab:s1_models}
        \end{table}

    \subsection{Preliminary Visualization}

\begin{figure}[ht]
            \centering
            \begin{tabular}{cc} 
            (a) SigLIP-SO400M & (b) DINOv2\\
            \includegraphics[width=0.47\linewidth]{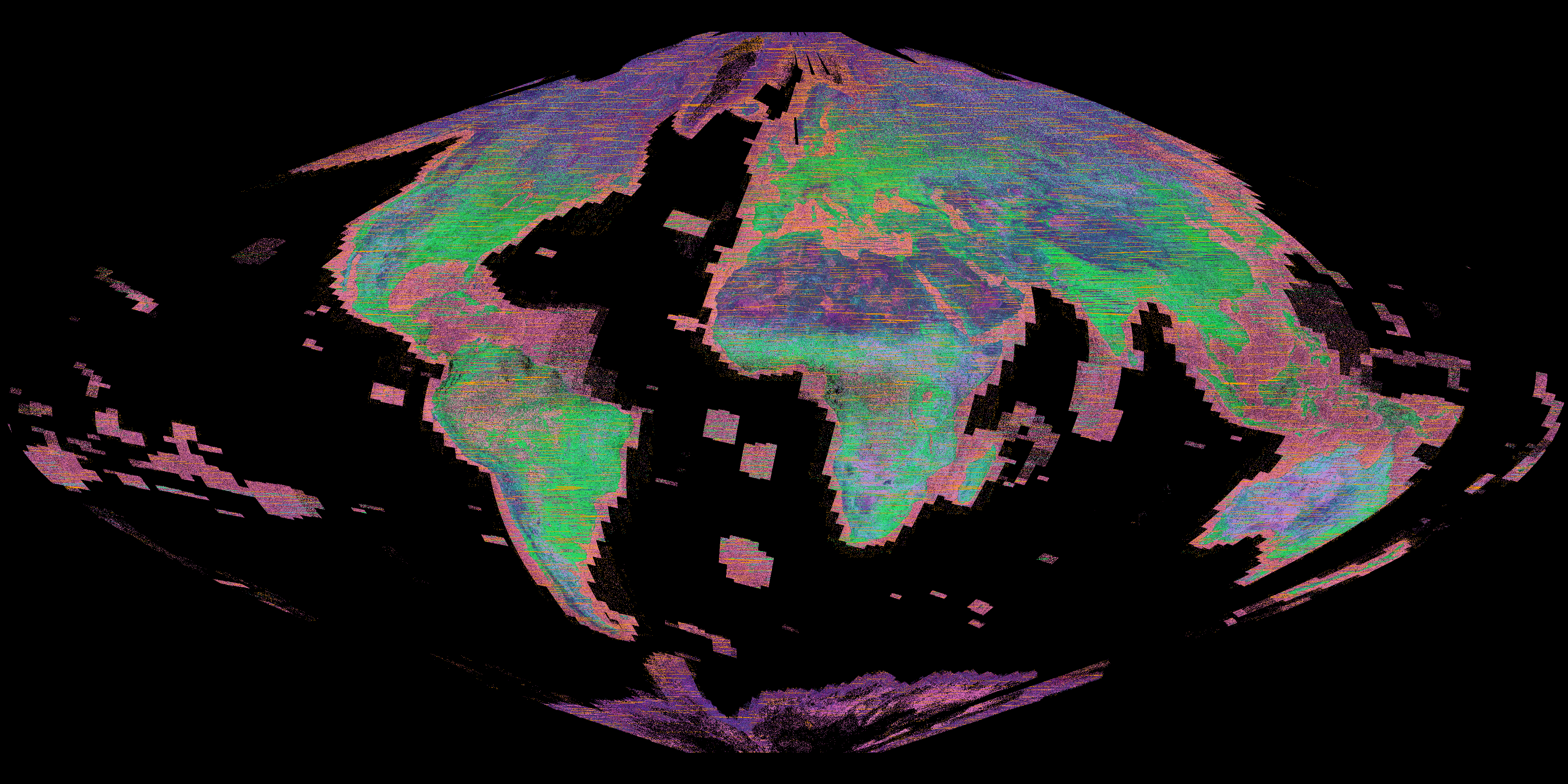} & 
            \includegraphics[width=0.47\linewidth]{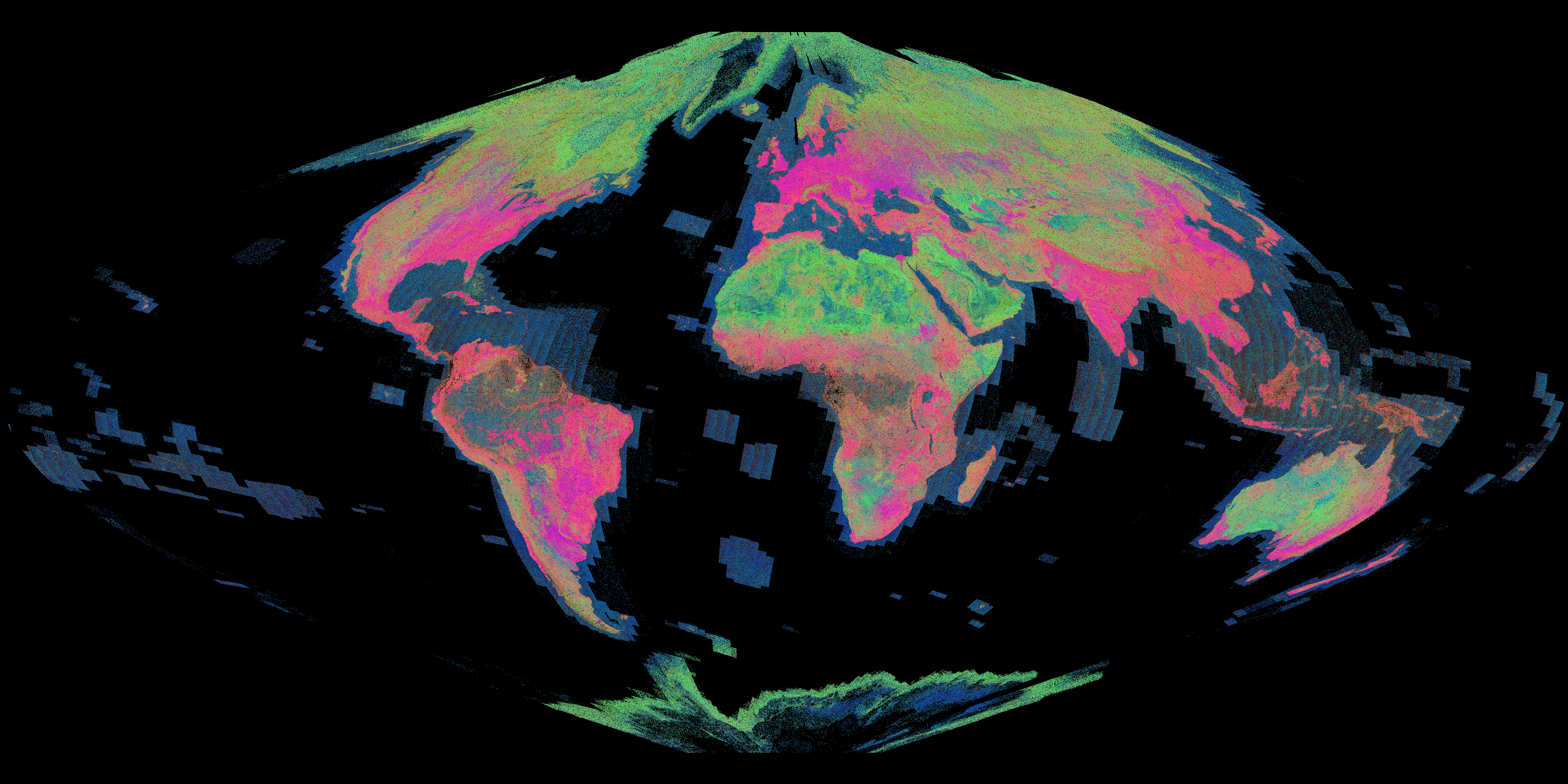} \\

            (c) SSL4EO-S2 & (d) SSL4EO-S1\\
            \includegraphics[width=0.47\linewidth]{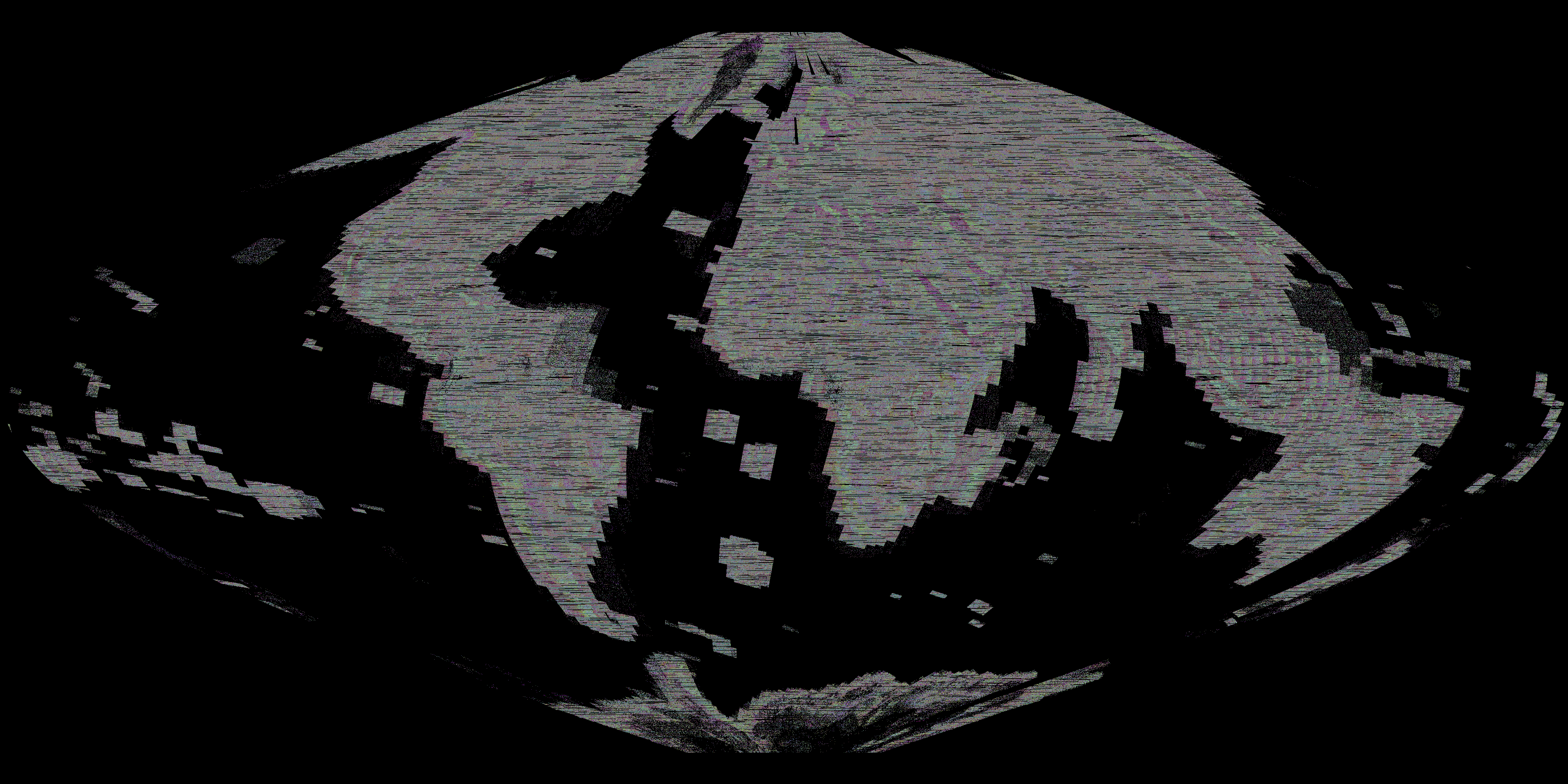}     & 
            \includegraphics[width=0.47\linewidth]{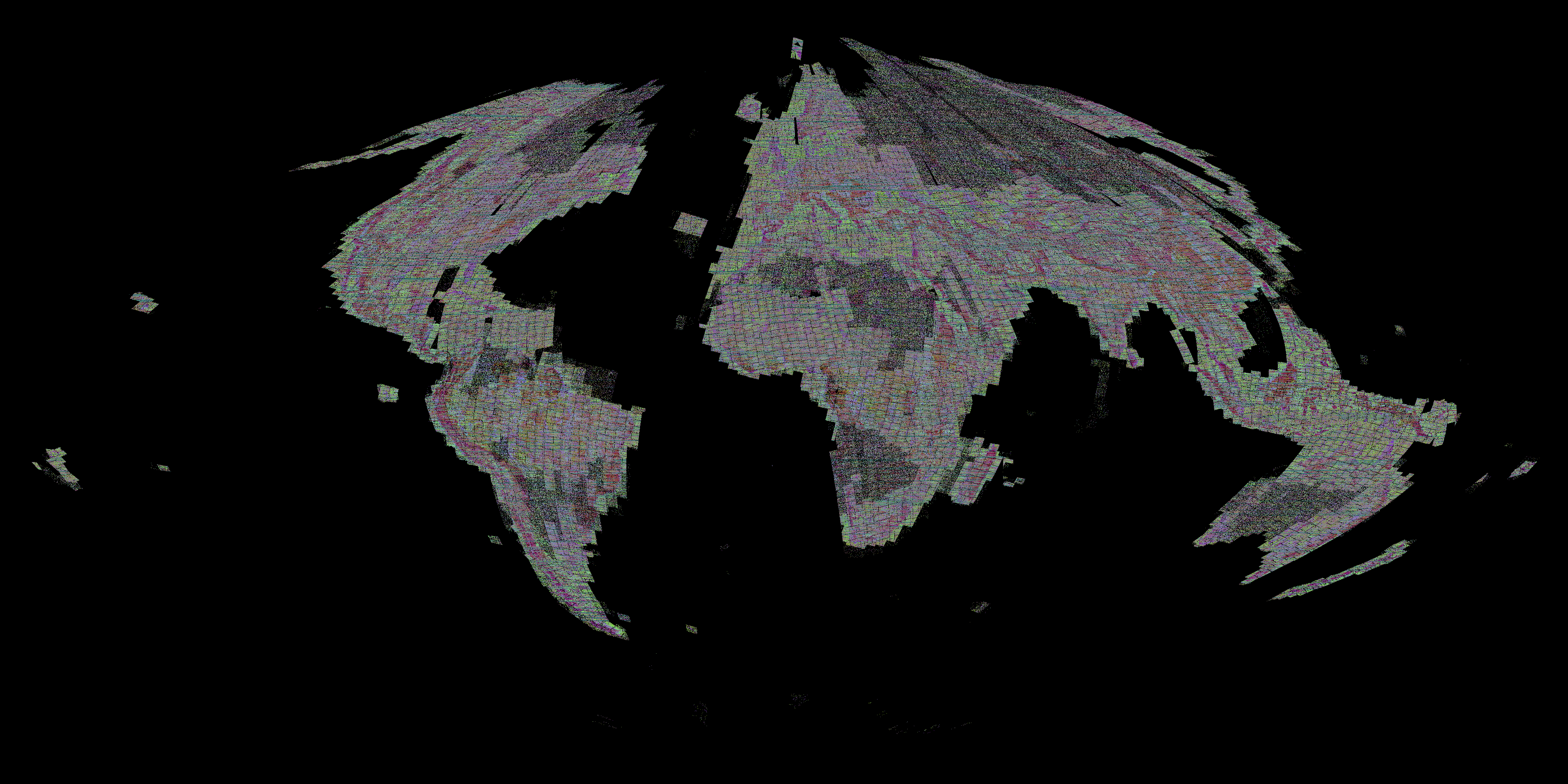}
            \end{tabular}
            \caption{Principal component analysis with 3 components mapped to RGB channels (larger format of the same images is available in Appendix~\ref{app:images}}
            \label{fig:pca_vis}
        \end{figure}

        Preliminary visualization via principal component analysis is shown in Figure~\ref{fig:pca_vis} for all 4 datasets. The results have been obtained by mapping the embedding from the central fragment of each Major TOM grid cell to a low-dimensional space of 3. The output vector of principal components is then mapped to the range of 8-bit unsigned integers for display, and the mean intensity is scaled to be the same for all three colors.

        While this is only an early step into the analysis process conducted immediately after producing the database, it can already shed some light into representation spaces encoded by individual models. Matching colors of two locations in the map indicate that the source images from those places possess aligned characteristic according to the embedding model.

        It is quite apparent that the embeddings produced by the general-purpose vision models (SigLIP and DINOv2) shown in Figure~\ref{fig:pca_vis}(a) and (b) appear to encode large areas into similar neighborhood of the latent space, exhibited by a shared colour offsets for large regions like norther Africa, the tropical belt, or the Alps. The embeddings from the SSL4EO models (Figure~\ref{fig:pca_vis}(c) and (d)) appear to be more conditioned on the local features, with less apparent global structure.

        These differences are only an introduction to a more detailed analysis, which is to be conducted in the upcoming work.
        
\section{Software Release}
\label{sec:software_release}

    There is already an existing codebase supporting the Major TOM project~\citep{majortom}. Since the embedding of Major TOM cells is now introduced as an expansion to the standard, the tools for generating and processing Major TOM embeddings are incorporated into the same package within the \href{https://github.com/ESA-PhiLab/Major-TOM/tree/main/src/embedder}{\texttt{src/embedder}} subdirectory.

    On the front end, it contains jupyter notebooks that allow the user to go through the complete process of transforming a Major TOM Core dataset into an embedding expansion. Furthermore, additional tools are provided for interacting with the embeddings, including visualization and operations on the vector database, with examples contained in other jupyter notebooks.

    \subsection{Embedding Generation}

        The embedding generation notebook covers the complete process of transforming each original data parquet to a parquet of embeddings, which involves the extraction of the relevant data bands for the embedding model, fragmentation function, preprocessing, model application, and finally, aggregation into output parquet data frames.

        Furthermore, in the upcoming release, additional tools for parallelizing the process are going to be released to improve the speed of the process for large-scale settings.

    \subsection{Embedding Interface}

        Once the embedding dataset has been built, there are many ways to interact with the vector database. The embedder subpackage of Major TOM contains examples of performing PCA at a global scale, performing similarity search, and optimising the memory required for those operations (for example, by reducing the precision of the number format).

    \subsection{Embedding Evaluation}

        This document describes the project at the time of the first release of the large-scale embedding data produced in the first stage of the undertaking. It is foreseen that with this ready access to millions of embeddings from several models, it may become possible to evaluate the capability of the learned representations, including downstream tasks.

        The software release is going to soon be expanded by a set of scripts for fast evaluation of the embedding models to make the process fast, reproducible, and expandable by the wider community.
        
\section{Potential Use Cases}
\label{sec:use_cases}

    Precomputed embedding datasets can serve concrete downstream use cases, such as land use monitoring, by learning a mapping from a given embedding space to label space corresponding to a given task. This reduces the need for expensive deep learning model inference, and can instead rely solely on computationally inexpensive mappings such as linear transformations.

    In the more general sense, ready and fast access to embeddings from several models allows to compare them next to each other across a number of tasks, regions, and times. By providing precomputed embeddings on benchmark datasets such as Major TOM, general-purpose models for Earth observation data become more reproducible, transparent, and accessible, since analysis can be conducted by a wider community of developers and users.

    The availability of large vector databases describing Earth observation data can provide a foundation for developing improved mechanisms for interacting with the embeddings, via operations such as similarity search or compression. The operations at the scale of millions of vectors become computationally demanding, and further research is required to scale these methods up to the level of Earth observation archives, which could easily contain trillions of vectors.
 
\vskip 0.2in

\newpage
\appendix
\section{Computational Budget}
\label{app:budget}
    The computational tasks in this study were carried out on the CloudFerro CREODIAS public cloud computing platform, utilizing OpenStack cloud technology. The hardware configuration included two GPUs (NVIDIA L40S with 48GB of VRAM), SSD disks were used for fast reading and writing, 64 VCPUs, and 240GB of RAM, running on Ubuntu 22.04. The environment was built on Python 3.9, with the deep learning framework Torch 2.4.1, utilizing GPU acceleration through CUDA 12.2. For image processing, OpenCV was employed for resampling operations, and multithreading libraries ensured optimized performance across multiple CPU cores. The TorchGeo library was used for interfacing with the pre-trained SSL4EO models. Major TOM Core datasets were downloaded directly from HuggingFace in the parquet format. The embedding scripts were set up via parallelized processes.

\section{Additional Images}
\label{app:images}

    \begin{figure}[ht]
        \centering
        \includegraphics[width=\linewidth]{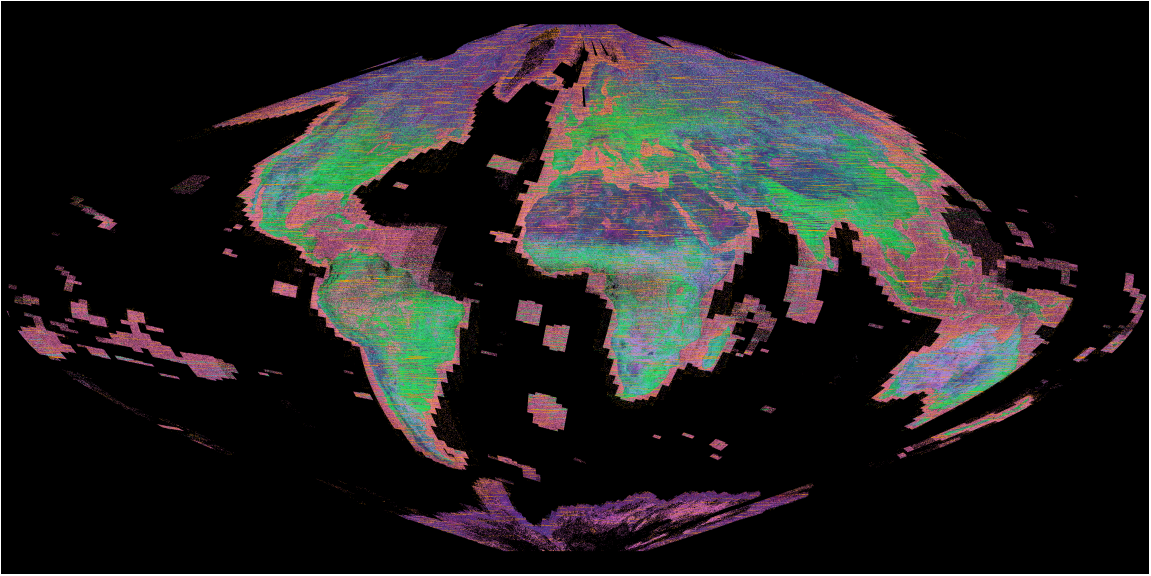}
        \caption{Principal component analysis with 3 components mapped to RGB channels for SigLIP-SO400M}
        \label{fig:pca_vis_siglip}
    \end{figure}

    \begin{figure}
        \centering
        \includegraphics[width=\linewidth]{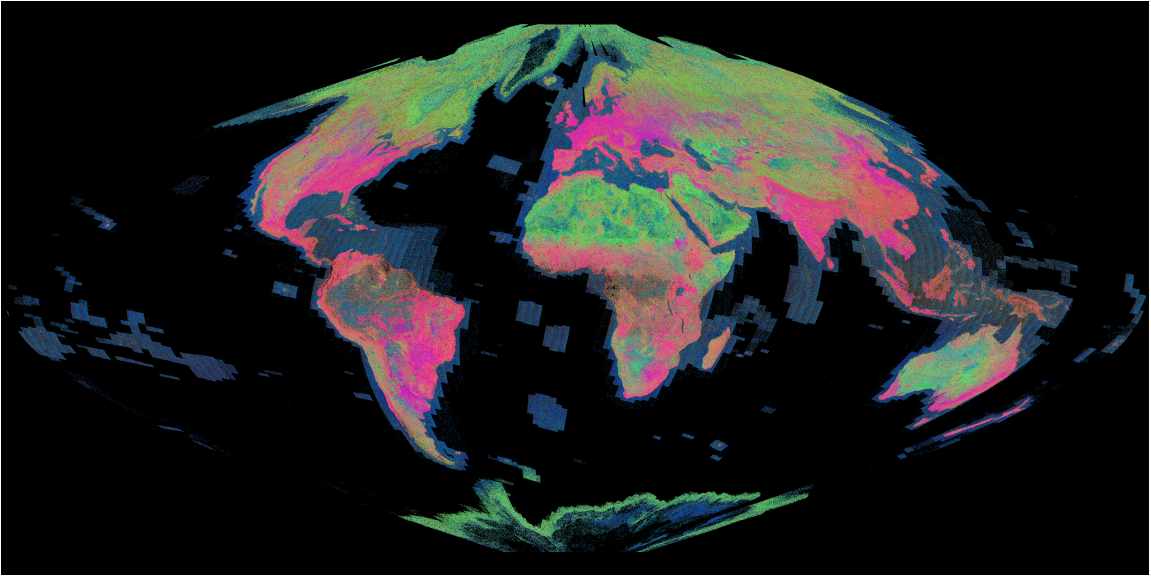}
        \caption{Principal component analysis with 3 components mapped to RGB channels for DINOv2}
        \label{fig:pca_vis_dino}
    \end{figure}

    \begin{figure}
        \centering
        \includegraphics[width=\linewidth]{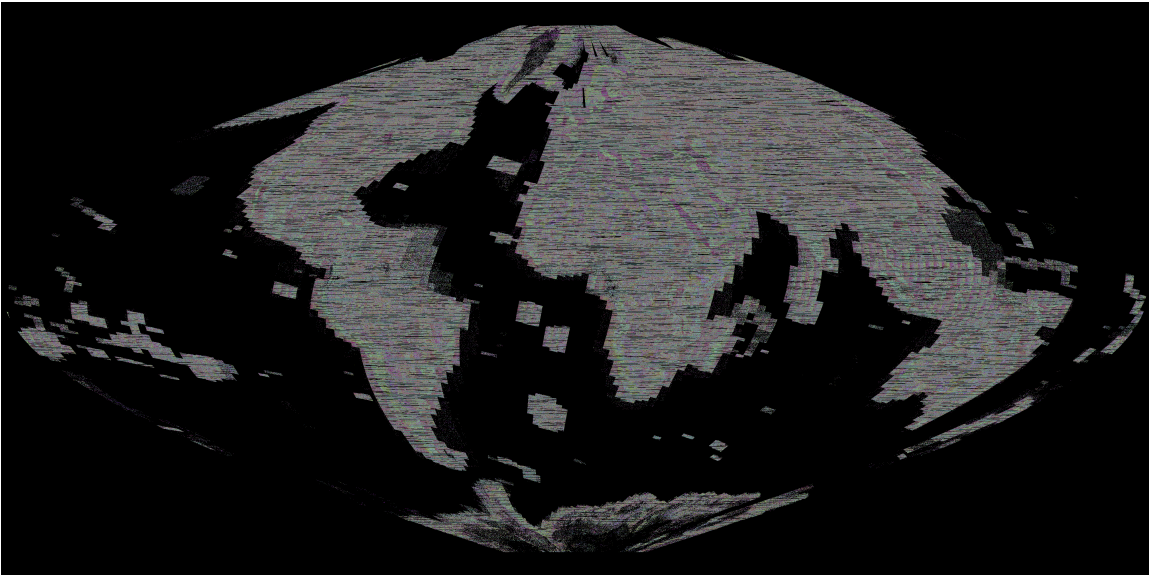}
        \caption{Principal component analysis with 3 components mapped to RGB channels for SSL4EO-S2}
        \label{fig:pca_vis_ssl4s2}
    \end{figure}
    
    \begin{figure}
        \centering
        \includegraphics[width=\linewidth]{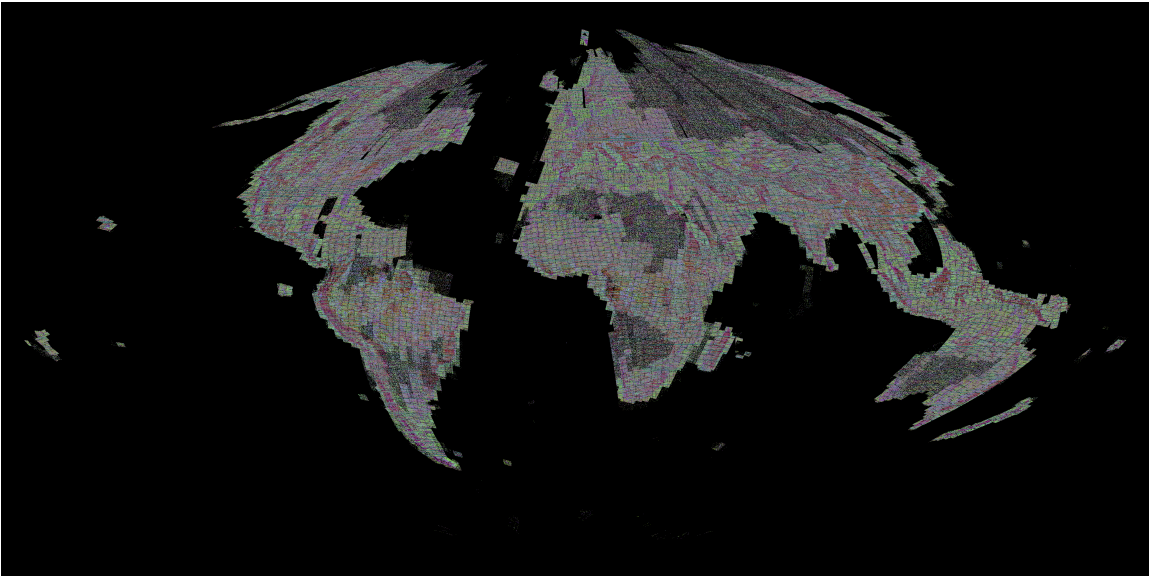}
        \caption{Principal component analysis with 3 components mapped to RGB channels for SSL4EO-S1}
        \label{fig:pca_vis_ssl1s1}
    \end{figure}

\end{document}